# EVARS-GPR: EVent-triggered Augmented Refitting of Gaussian Process Regression for Seasonal Data

Florian Haselbeck[1,2,[0000-0002-5702-376X]] and Dominik G. Grimm[1,2,3,[0000-0003-2085-4591],(✉)]

[1] Technical University of Munich, TUM Campus Straubing for Biotechnology and Sustainability, Bioinformatics, Schulgasse 22, 94315 Straubing, Germany
[2] Weihenstephan-Triesdorf University of Applied Sciences, Bioinformatics, Petersgasse 18, 94315 Straubing, Germany
[3] Technical University of Munich, Department of Informatics, Boltzmannstr. 3, 85748 Garching, Germany
{florian.haselbeck, dominik.grimm}@hswt.de

**Abstract.** Time series forecasting is a growing domain with diverse applications. However, changes of the system behavior over time due to internal or external influences are challenging. Therefore, predictions of a previously learned forecasting model might not be useful anymore. In this paper, we present **EV**ent-triggered **A**ugmented **R**efitting of Gaussian Process Regression for **S**easonal Data (EVARS-GPR), a novel online algorithm that is able to handle sudden shifts in the target variable scale of seasonal data. For this purpose, EVARS-GPR combines online change point detection with a refitting of the prediction model using data augmentation for samples prior to a change point. Our experiments on simulated data show that EVARS-GPR is applicable for a wide range of output scale changes. EVARS-GPR has on average a 20.8 % lower RMSE on different real-world datasets compared to methods with a similar computational resource consumption. Furthermore, we show that our algorithm leads to a six-fold reduction of the averaged runtime in relation to all comparison partners with a periodical refitting strategy. In summary, we present a computationally efficient online forecasting algorithm for seasonal time series with changes of the target variable scale and demonstrate its functionality on simulated as well as real-world data. All code is publicly available on GitHub: https://github.com/grimmlab/evars-gpr.

## 1 Introduction

Time series forecasting is an emerging topic with applications in diverse domains, e.g. business, medicine or energy. These approaches make use of time series data, which describes a system behavior by a sequence of observations within a certain time period and try to predict future values. However, sudden changes of the system behavior over time are common issues in time series analysis. These sudden changes can be either caused by external or internal influences, e.g. due to operational or strategic decisions.

For instance, currently many sales forecasting systems are affected by the SARS-CoV-2 pandemic and energy demand predictions might be impeded by energetic optimizations of big consumers. Some but probably not all of the influential factors can be captured by features. Nevertheless, after a change of the generative data distribution, which reflects the relation between explanatory features and the target variable, predictions of a previously learned model might not be useful anymore. As a result, decisions based on these could cause damage such as a financial loss if e.g. an underestimated demand leads to missed sales [2, 13].

A common but computationally exhaustive approach to handle this problem is to periodically retrain a prediction model during the productive operation [13]. Furthermore, several methods combine change point detection (CPD), i.e. the problem of identifying a change of the generative data distribution, and Gaussian Process Regression (GPR). Some of them work offline and are therefore not suitable for changing data distributions during the online phase [9, 16]. Existing online approaches are either not event-triggered [21, 22], require a certain number of samples of a new generative distribution [12] or are based on *a priori* assumptions in terms of potentially changing time series properties [17]. Furthermore, none of them apply data augmentation (DA) on samples prior to a detected change point to reuse these augmented samples for model retraining.

In this work, we present **EV**ent-triggered **A**ugmented **R**efitting of Gaussian Process Regression for **S**easonal Data (EVARS-GPR), for which we provide an overview in Fig. 1. This novel online algorithm combines change point monitoring with a refitting of the prediction model using data augmentation. Compared to existing approaches, the main focus of our algorithm is on seasonal data with sudden changes of the target variable scale while values of explanatory features remain approximately equal, which is a common issue in seasonal time series forecasting. The data augmentation step is triggered after the detection of a change point and a deviation of the target variable scale compared to a certain threshold. This step updates known samples prior to a change point with new information on the changed target variable scale. Consequently, we gain potential useful data for the refitting of the prediction model. Hence, EVARS-GPR is event-triggered and as a result more efficient than a periodical refitting strategy. Furthermore, the algorithm reacts immediately after a detected change point and *a priori* assumptions on the output scale changes are not required. As prediction model we use Gaussian Process Regression (GPR), see Appendix 1 for an overview. GPR is a flexible and non-parametric Bayesian method including uncertainties of a prediction value, which seems profitable with regard to the practical use of forecasts. Moreover, we evaluate the integration of different approaches for online CPD and DA, two essential parts of EVARS-GPR. We further analyze EVARS-GPR using simulated data and evaluate the performance on real-world datasets including different comparison partners.

The remainder of this paper is organized as follows. In section 2, we describe the related work. Afterwards, we provide the problem formulation in section 3. Then, we outline EVARS-GPR in section 4 followed by the experimental setup in section 5. The experimental results are shown and discussed in section 6, before we draw conclusions.

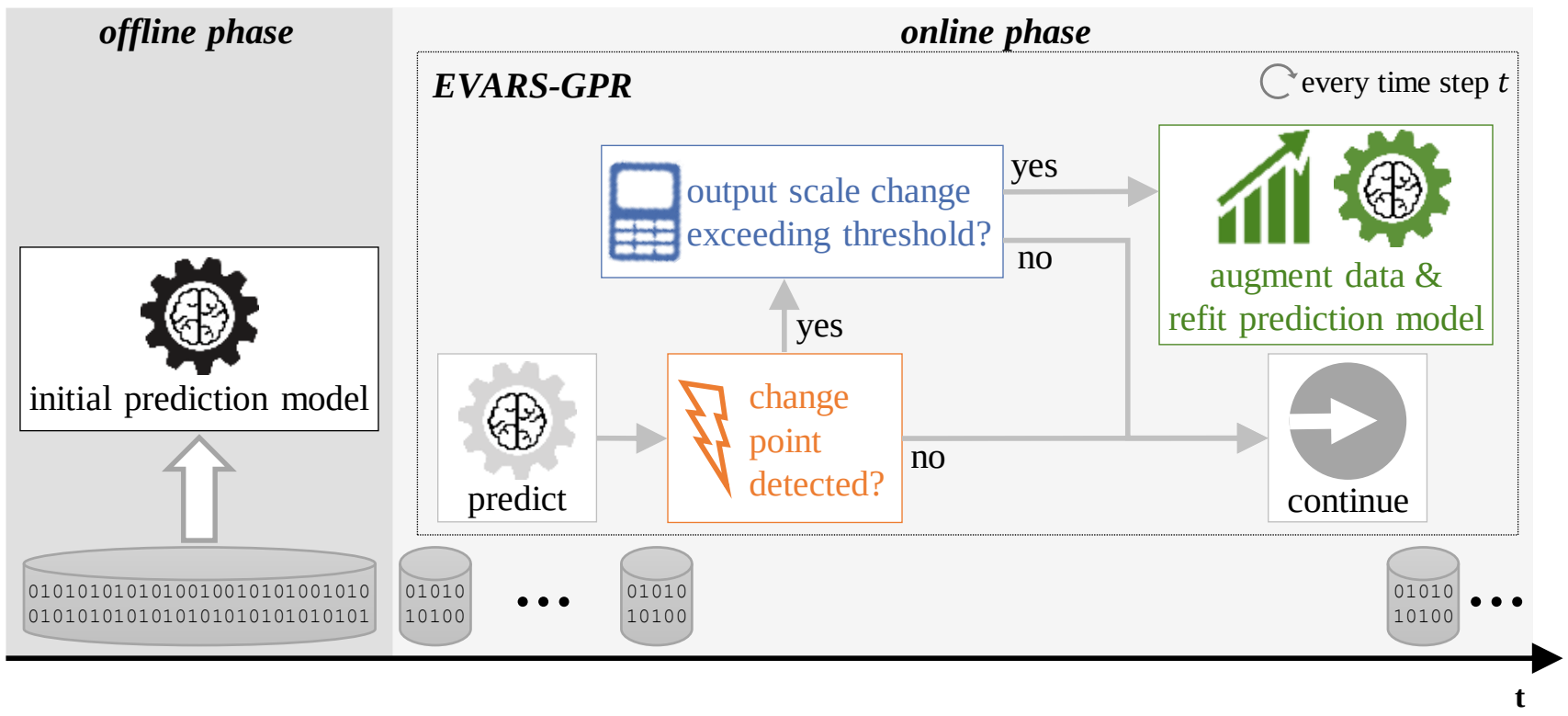

**Fig. 1. Overview of EVARS-GPR during the online phase and the preconditions in the offline phase.** The initial prediction model is trained offline. During the online phase, the prediction of the next target value is followed by a change point detection. If a change point is detected, the output scaling factor, which sets the target values of the current season in relation to previous seasons, is calculated. If the deviation between the current and last output scaling factor exceeds a threshold, then an augmented refitting of the prediction model is triggered. In case one of the two conditions is not fulfilled, EVARS-GPR continues using the current prediction model.

## 2 Related Work

Methods enabling GPR models to work with nonstationary data distributions can be divided into offline and online techniques. A common offline approach is to switch between kernel functions, e.g. by multiplication with sigmoid functions [9]. For some technical processes, multiple steady states can be determined. This enables the association of a corresponding model. For inference, the one associated with the current state is selected [16]. However, these approaches are limited to scenarios for which change points respectively steady states can be defined *a priori*. Furthermore, change points occurring abruptly at a single point can be treated as a hyperparameter of a nonstationary covariance function [11]. A further possibility of handling nonstationary data distributions is to augment the input using time-related functions. One option is the introduction of a forgetting factor, which leads to a lower influence of the information contained in older samples. Another common technique is to periodically update the hyperparameters of a GPR model using samples within a specified moving window [22]. A more elaborate approach is Moving-Window GPR (MWGPR). This method discards the oldest sample after a new one becomes available. Moreover, a dual preprocessing and dual updating strategy is performed. This introduces a recursive bias term, which depends on past model errors and is added to the model's prediction to get the final forecast value [21]. All these approaches have the drawback of losing potentially useful information from earlier samples even if the data distribution did not change [22]. GP-non-Bayesian clustering (GP-NBC) is focused on computational efficiency with the goal of making it suitable for resource-constrained environments, e.g. robotic platforms. Based on online-trained GP models, likelihood ratio tests are performed in order to determine whether a new candidate model or a previously stored one should be used in the further process. A disadvantage of GP-NBC is that a certain number of new samples needs to be available to enable the training of a new model. This may lead to a delayed reaction after a change point occurred [12]. The INstant TEmporal structure Learning (INTEL) algorithm was recently proposed. First, a template model is learned using offline data. Then, a set of candidate models with varying hyperparameters due to assumptions of potential changes during the productive operation is constructed based on the template model. For the final prediction, all models are combined using weights that correspond to the likelihood of a new observation given each model. In its current implementation, INTEL is limited to univariate data. Furthermore, possible changes happening during the online phase need to be assumed *a priori* [17].

## 3 Problem Formulation

We define a multivariate time series $\mathcal{D} = \{\boldsymbol{x}_t, y_t\}^n$ as a sequence of $n$ samples consisting of $d$-dimensional explanatory variables called covariates $\boldsymbol{x}_t \in \mathbb{R}^d$ and a corresponding target value $y_t \in \mathbb{R}$ at time step $t$. The target value at time step t, $y_t$, is drawn from a distribution $p_i(y|\boldsymbol{x}_t)$, i.e. it is dependent on the covariates $\boldsymbol{x}_t$. In this work, we consider seasonal data, meaning data that follows a certain periodicity of length $n_{seas}$. We assume that periodicity is present for the target variable $y$ as well as for at least some of the covariates $\boldsymbol{X}$. Thus, the target variable at time step $t$ can be decomposed in a seasonal component $s_t$ and a residual $r_t$ summing up all other effects: $y_t = s_t + r_t$. Based on its periodicity of length $n_{seas}$, the seasonal component at time step t is similar to those of previous seasons: $s_t \approx s_{t-k \cdot n_{seas}}$ with $k \in \mathbb{N}\backslash\{0\}$. The covariates $\boldsymbol{x}_t$ respectively a subset $\boldsymbol{\chi}_t \subseteq \boldsymbol{x}_t$ of them can also be decomposed in a seasonal component $\boldsymbol{s}_{\chi,t}$ and a residual $\boldsymbol{r}_{\chi,t}$ into $\boldsymbol{\chi}_t = \boldsymbol{s}_{\chi,t} + \boldsymbol{r}_{\chi,t}$, with similar periodicity characteristics regarding $\boldsymbol{s}_{\chi,t}$. The strength of the seasonal pattern, i.e. the influence of the seasonal component on the final value, might vary for different covariates and target variables.

With $n_{off}$ samples of $\mathcal{D}$, a model $M$, here a Gaussian Process Regression, can be trained offline using cross-validation to determine the hyperparameter configuration that delivers predictions $\hat{y}$ generalizing best to the true distribution $p_i(y|\boldsymbol{x})$. During the online phase, with a new input $\boldsymbol{x}_t$ provided at every time step $t$, the model $M$ is used to deliver a prediction for the target variable value $\hat{y}_t$ based on $\boldsymbol{x}_t$. However, it is a common issue in time series forecasting that the generative distribution $p_i(y|\boldsymbol{x})$ our predictor $M$ was trained on might change to another distribution $p_j(y|\boldsymbol{x})$. The time step at which such a shift happens is called a change point. In this work, we focus on output scale shifts, meaning that the value range of the target variable $y$ changes. Therefore, with regard to the periodicity of the covariates $\boldsymbol{X}$ and the target variable $\mathbf{y}$, a similar covariate vector $\boldsymbol{x}_t$ corresponds to a different target variable $y_t$ as the generative distribution changed. Consequently, the predictions produced by the previously trained model $M$ might not be useful anymore. With EVARS-GPR, we address this problem by combining online change point monitoring of the target variable $y$ and a refitting of the base model $M$ using data augmentation in case a change point is detected. A list of symbols including those of subsequent sections is provided in Appendix 2.

# 4 EVARS-GPR

EVARS-GPR is an online algorithm that is focused on changes resulting in an output scale shift of seasonal multivariate time series, as outlined in section 3. In Fig. 1 and Algorithm 1, we give an overview of EVARS-GPR. Following the problem formulation, we assume an offline-trained model $M$, which we subsequently call the base model $M_{base}$. Prior to the online phase, the current prediction model $M_{current}$ is initialized with this offline-trained model $M_{base}$. As EVARS-GPR operates online, the main part starts with a new sample becoming available at time step $t$. As a first step, we retrieve the prediction of the next target value $\hat{y}_t$ using the covariates $\boldsymbol{x}_t$ as well as the current model $M_{current}$. Then, we run an online change point detection (CPD) algorithm, which is updated with the current target variable value $y_t$. In case we do not detect a shift of the generative distribution $p(y|\boldsymbol{x}_t)$, the current prediction model $M_{current}$ stays unchanged and the algorithm waits for the next time step $t+1$. However, if we determine a change point at time step $t$, the remaining procedures of EVARS-GPR are triggered. First, as EVARS-GPR is focused on changes of the output scale in seasonal time series data, the output scaling factor $\eta$ is determined. For that purpose, the target values $y$ prior to the change point and within a window of size $n_w$ are considered. These are set in relation to the target values $y$ within the corresponding window of $n_\eta$ previous seasons with a season length of $n_{seas}$:

$$\eta = \frac{1}{n_\eta} \sum_{k=1}^{n_\eta} \frac{\sum_{i=t-n_w}^{t} y_i}{\sum_{j=t-k\cdot n_{seas}-n_w}^{t-k\cdot n_{seas}} y_j} \tag{1}$$

The nominator of Eq. (1) includes current target values $y_i$ prior to the change point, whereas the denominator conveys information on the corresponding period of a previous season. This ratio is averaged over the number of seasons taken into account to retrieve the output scaling factor. Online CPD is prone to false alarms due to outliers. For this reason and to limit the amount of refittings for efficiency, we set a minimum threshold $\pi_\eta$ for the deviation between the current output scaling factor $\eta$ and the output scaling factor of the last augmented refitting $\eta_{old}$. If this threshold is exceeded, the augmented refitting of the current model $M_{current}$ is triggered. First, we generate an augmented set of samples $\mathcal{D}'$ based on the dataset prior to the change point at time step $t$. Thereby, we reuse known samples and update them with new information on the changed target variable scale. Consequently, we gain an augmented dataset $\mathcal{D}'$ for the refitting of the current model $M_{current}$. Furthermore, the last output scaling factor $\eta_{old}$ is stored. Subsequently, the refitted current model is used for the predictions of the target value. With a new sample arriving at the next time step $t+1$, the whole cycle of predicting, change point monitoring, potential data augmentation and model refitting starts again.

The goal of CPD is to find abrupt changes in data, in the context of this work resulting in a shift of the scale of the target variable $y$. A CPD method should ensure a quick reaction to a change point. Considering a real time operation, computationally efficient

CPD methods are advantageous. Beyond that, for EVARS-GPR, the CPD and the prediction methods are separated in order to enable the output scale-dependent, augmented model refitting. For these reasons, we excluded approaches such as GPTS-CP [24] and BOCPD-MS [15]. Based on the outlined criteria, we evaluated Bayesian Online Change Point Detection (BOCPD) and ChangeFinder (CF). More information on these two methods can be found in Appendix 3. In both cases, we deseasonalize data via seasonal differencing in order to prevent false alarms due to seasonal effects [2].

Besides online CPD, DA is an essential part of EVARS-GPR. For this work, we focus on computationally efficient approaches ensuring a real time operation and consider small datasets as well. Therefore, we excluded generative models such as TimeGAN [28] or C-RNN-GAN [20]. First, we augmented the original dataset consisting of all samples prior to a change point at time step $t$, $\mathcal{D}_{0:t}$, by scaling the original target variable vector $\boldsymbol{y}_{0:t}$. Thereby, we multiply the target variable vector $\boldsymbol{y}_{0:t}$ with the output scaling factor $\eta$ and leave the covariates $\boldsymbol{x}_{0:t}$ unchanged, resulting in the augmented dataset $\mathcal{D}^{\eta}_{0:t}$. Considering the focus on shifts of the output scale, augmenting the dataset by scaling the target variable vector $\boldsymbol{y}$ is a reasonable and efficient approach. Second, we used two virtual sample generation techniques for imbalanced regression: Random Oversampling with the introduction of Gaussian Noise (GN) [26] and SMOGN, which combines the former and the Synthetic Minority Oversampling TEchnique for Regression (SMOTER) [5, 27]. Both methods are outlined in Appendix 4.

**Algorithm 1.** EVARS-GPR

**Inputs:** $M_{base}$, $\mathcal{D}_{0:n_{off}}$
**Parameters:** $n_\eta, n_{seas}, n_w, \pi_\eta$, *CPD* and *DA* parameters
**Results:** $\hat{y}$

$M_{current} = M_{base}$
$\eta_{old} = 1$
**for** *new sample at time step t* **do**
    predict target value: $\hat{y}_t = M_{current}(\boldsymbol{x}_t)$
    perform online CPD: $online_cpd(\boldsymbol{y}_{0:t}, CPD\ parameters)$ ▷ App. 2
    **if** *change point detected* **then**
        calculate output scaling factor: $\eta = calc_output_scaling_factor(n_\eta, n_{seas}, n_w)$ ▷ Eq. (1)
        **if** $|\eta - \eta_{old}| \,/\, \eta_{old} > \pi_\eta$ **then**
            augment data: $\mathcal{D}' = augment_data(\mathcal{D}_{0:t}, DA\ parameters)$ ▷ App. 3
            refit current model: $refit(M_{current}, \mathcal{D}')$
            $\eta_{old} = \eta$
        **end if**
    **end if**
**end for**

# 5 Experimental Setup

Subsequently, we will first describe the simulated data we used to determine the configuration of EVARS-GPR and to analyze its behavior. Afterwards, we outline the real-world datasets and the performance evaluation.

## 5.1 Simulated Data

EVARS-GPR is focused on seasonal data with changes regarding the target variable scale during the online phase. In order to configure and parametrize the algorithm as well as to analyze its behavior, we generated simulated data fulfilling these properties. Fig. 2 visualizes a simplified example of simulated data to explain its configuration. As we observe in Fig. 2a, the target variable $\boldsymbol{y}$ follows a periodical pattern with a season length $n_{seas}$ and an amplitude $a$. Between $t_{start}$ and $t_{end}$ during the online phase, we manipulate $\boldsymbol{y}$ by a multiplication with a factor $\delta$, which results in a change of the output scale. The characteristics of this manipulation factor are visualized in Fig. 2b. At $t_{start}$, we begin with $\delta_{base}$ and increase $\delta$ by a slope of $\kappa$ at every time step $t$, up to a maximum manipulation factor $\delta_{max}$. Then, the manipulation factor $\delta$ stays constant until its sequential decrease is triggered in order to reach $\delta_{base}$ at $t_{end}$. To meet the properties specified in section 3, the covariates $\boldsymbol{x}$ are also periodical. Both $\boldsymbol{x}$ and $\boldsymbol{y}$ can be modified with additive random noise in order to model the seasonality more realistic.

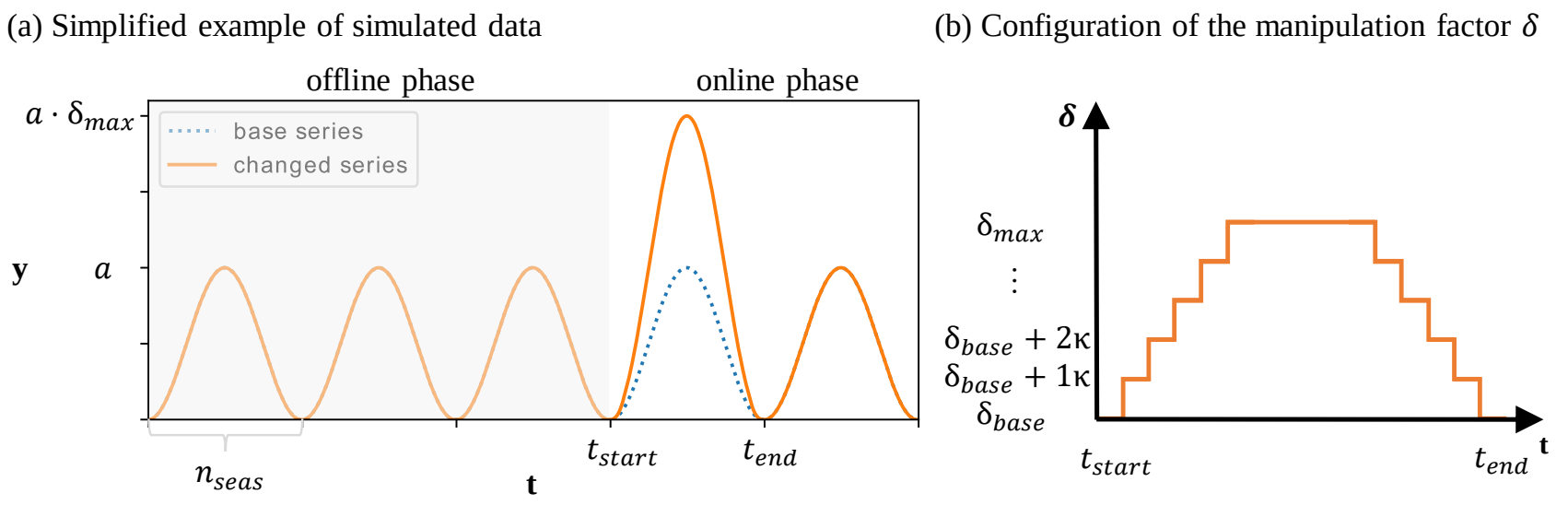

**Fig. 2. (a) Visualization of a simplified example of simulated data** with the base as well as the changed series, both with a season length of $n_{seas}$. Between $t_{start}$ and $t_{end}$, the base series with its amplitude $a$ is changed by multiplication with a manipulation factor $\delta$. **(b) Configuration of the manipulation factor $\boldsymbol{\delta}$.** Starting from $\delta_{base}$ at $t_{start}$, the manipulation factor $\delta$ increases by a slope $\kappa$ at every time step $t$. If a maximum manipulation factor $\delta_{max}$ is reached, $\delta$ stays constant. At $t_{end}$, the base factor $\delta_{base}$ is reached again after sequentially decreasing $\delta$ using $\kappa$.

In summary, the parameters $n_{seas}$, $t_{start}$, $t_{end}$, $\delta_{max}$ and $\kappa$ enable us to simulate various settings of the output scale change. For instance, $t_{start}$ and $t_{end}$ modify the duration and time of occurrence. Furthermore, $\delta_{max}$ marks the maximum extent of the output scale change, whereas $\kappa$ determines its increase at every time step $t$, thus the speed respectively abruptness. Based on this, we formulated 67 scenarios and evaluated the performance to select the online CPD and DA method for EVARS-GPR, see section 6.1. Furthermore, the parametrization of EVARS-GPR is based on these scenarios, see Appendix 5 for an overview. For that purpose, we employed a random search with 100 different parameter settings for each combination of online CPD and DA method [3].

### 5.2 Real-World Datasets

We additionally evaluated EVARS-GPR on real-world datasets. Based on the algorithm's scope, we selected seasonal time series data, for which we provide more information in Appendix 6. For the horticultural sales prediction dataset *CashierData*[1], we observe a strong sales increase of potted plants (*PotTotal*) during the SARS-CoV-2 pandemic in 2020. Furthermore, we included the following common and publicly available datasets with changes of the output scale during the online phase: *DrugSales [13], VisitorNights [13], AirPassengers [4]* and *MaunaLoa [10]*. Beyond that, we used time series data without such changes to test the robustness of EVARS-GPR: *ChampagneSales [18], TouristsIndia [7], Milk [19], Beer [13]* and *USDeaths [19]*. We further applied mean imputation for missing values and added calendric as well as statistical features, e.g. lagged target variables, see Appendix 6 for an overview. Then, we used 80 % of the data to determine the base model $M_{base}$, i.e. the configuration that leads to the best performance in a cross-validation setup. Thereby, we employed a random search over the model's hyperparameters such as the kernel function as well as preprocessing parameters, e.g. whether to perform a principal component analysis [3]. Finally, we evaluated EVARS-GPR in an online setting for the remaining left out 20 % of the data.

### 5.3 Evaluation

To evaluate the performance on a set of $n$ samples, we used the Root Mean Squared Error (RMSE), which is defined as $RMSE = \sqrt{1/\mathrm{n} \sum_{i=1}^{n}(y_i - \hat{y}_i)^2}$ with the true value $y_i$ and the prediction $\hat{y}_i$. As the RMSE is scale-dependent, we further applied a scaling by the RMSE value achieved with $M_{base}$, subsequently called RMSE-ratio. Thus, the performances on simulated scenarios with different scales are comparable.

We included several comparison partners for the real-world datasets. $M_{base}$ applies the offline-trained model during the whole online phase. Furthermore, we employed common but computationally exhaustive periodical refits of the prediction model, which trigger a retraining at every (*PR1*) respectively every second (*PR2*) time step [13]. Moving-Window GPR (MWGPR) is included as an additional computational resource demanding comparison partner, because a refit is needed at every time step $t$ [21]. Moreover, we defined methods with a computational resource consumption similar to EVARS-GPR. These methods also react after a valid change point was detected, so the number of refittings and thus the resource consumption is similar. *CPD_scaled* scales the predictions of $M_{base}$ using the output scaling factor $\eta$, whereas *CPD_retrain* triggers a refitting of the current prediction model $M_{current}$ using all original samples prior to a change point. *CPD_MW* also leads to a refitting, but only uses data of the season before the detected change point. For an estimate of the resource consumption, we measured the process-wide CPU time of EVARS-GPR and the computationally exhaustive methods on a machine with two 2.1 GHz *Intel Xeon Gold 6230R* CPUs (each with 26 cores and 52 threads) and a total of 756 GB of memory.

---

[1] https://github.com/grimmlab/HorticulturalSalesPredictions

# 6 Experimental Results

In this section, we will first describe the behavior on simulated data. Afterwards, we outline the results on real-world datasets, before we discuss them.

## 6.1 Behavior on Simulated Data

We determined the configuration and parameters of EVARS-GPR based on simulated data. Using CF for online CPD and a scaling of the original dataset for DA lead to the lowest RMSE-ratio averaged over all scenarios (0.549). Thereby, we experienced that EVARS-GPR is sensitive to hyperparameters, e.g. the window size $n_w$, see Appendix 5 regarding the final values. We further analyzed EVARS-GPR on different output scale changes regarding the extent (maximum manipulation factor $\delta_{max}$), speed (slope $\kappa$), time of occurrence and duration (both via start $t_{start}$ respectively end index $t_{end}$). Results are visualized in Fig. 3. The performance of EVARS-GPR was in all scenarios at least equal to $M_{base}$ and outperformed it in most of the cases. Fig. 3a and 3b show that the advantage of EVARS-GPR was larger for longer periods with an output change. However, there was also an improvement for shorter durations. EVARS-GPR was beneficial for several extents and speeds of the shift as well as robust for constant scenarios ($\delta_{max} = 1$), see Fig. 3c and 3d. For smaller slopes $\kappa$, the improvement tended to decrease with a higher maximum manipulation factor $\delta_{max}$. We observe in Fig. 3d that EVARS-GPR's benefit was mostly smaller for maximum manipulation factors $d_{max}$ close to one. We included further scenarios in Appendix 7, which show similar results.

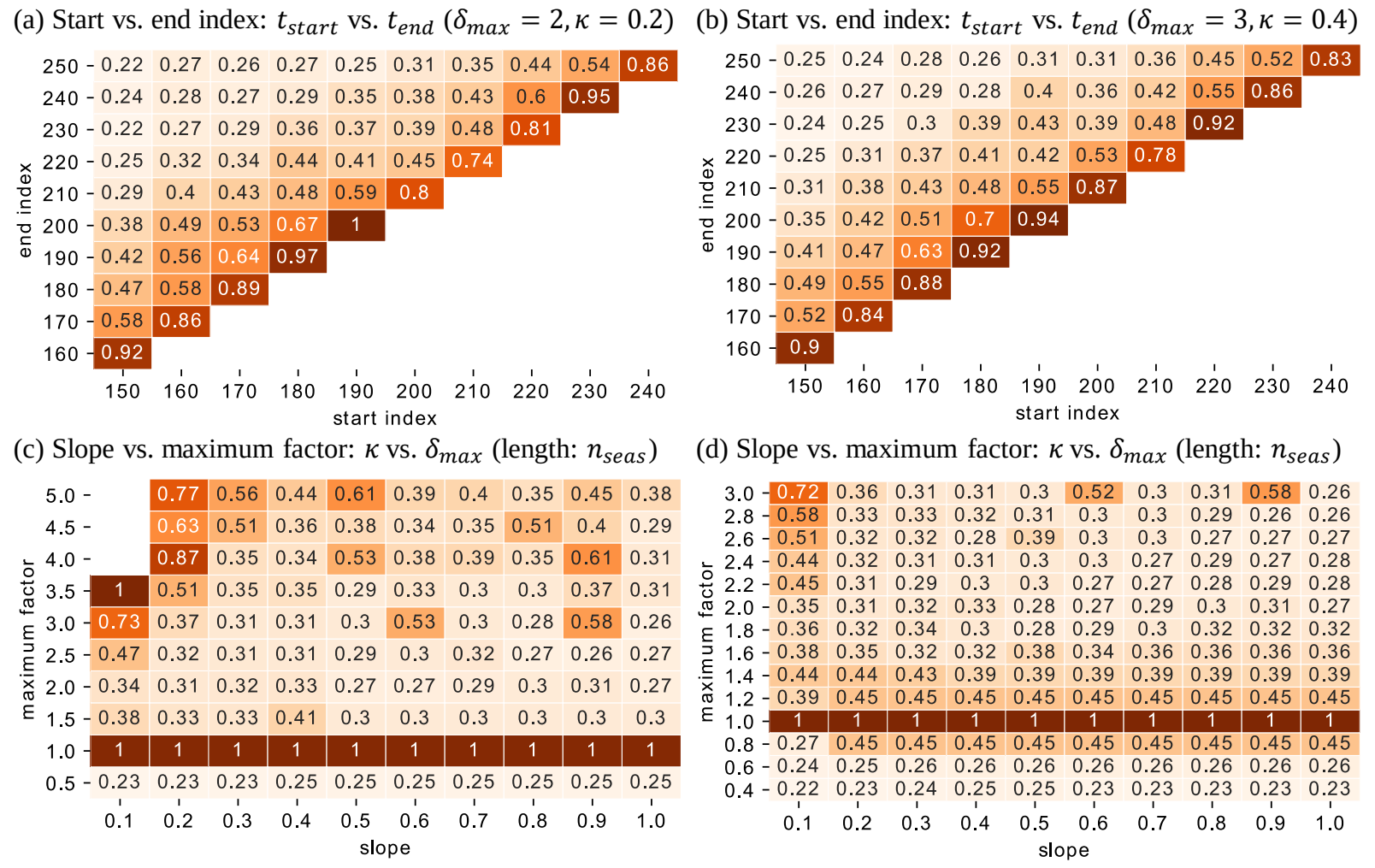

**Fig. 3. Behavior on a variety of simulated data ($n_{seas} = 50$).** Each box shows the result of the scenario parametrized with the values given on the x- and y- axis. Numbers are RMSE-ratios, lower values reflect a higher improvement compared to $M_{base}$. We analyzed the following factors of the output scale change: **(a)** time of occurrence and duration via start and end indices $t_{start}$ respectively $t_{end}$, **(b)** same factors with a higher slope $\kappa$ and maximum manipulation factor $\delta_{\max}$, **(c)** extent and speed of the change via $\kappa$ and $\delta_{max}$, **(d)** same factors on a finer grid for $\delta_{max}$.

### 6.2 Results on Real-World Datasets

We further evaluated EVARS-GPR on several real-world datasets. In Table 1, we show the model performance in terms of RMSE compared to methods with a similar computational resource consumption for datasets with a changing target variable scale. As we observe, EVARS-GPR was superior on all datasets. Our algorithm outperformed $M_{base}$ with an improvement of 37.9 % averaged over all datasets and the second-best competing method among all datasets by 20.8 %.

**Table 1. Performance comparison based on RMSE for datasets with output scale changes during the online phase for EVARS-GPR and methods with a similar resource consumption.** Numbers show the RMSE of the simulated online phase. The best results are printed bold.

| | $M_{base}$ | *CPD_scaled* | *CPD_retrain* | *CPD_MW* | *EVARS-GPR* |
|---|---|---|---|---|---|
| *CashierData* | 1351.43 | 2266.57 | 1314.55 | 1683.11 | **1125.34** |
| *DrugSales* | 6.15 | 5.39 | 3.90 | 4.46 | **2.75** |
| *AirPassengers* | 171.58 | 101.61 | 108.79 | 174.31 | **93.88** |
| *MaunaLoa* | 34.37 | 32.01 | 31.22 | 33.50 | **27.96** |
| *VisitorNights* | 10.97 | 9.30 | 8.80 | 10.34 | **5.11** |

Beyond that, we compared EVARS-GPR to computationally exhaustive methods, for which we show the results as well as the process-wide CPU time in Table 2. EVARS-GPR outperformed all other methods with respect to the CPU time. In comparison to *PR2*, the method with the second lowest runtimes, the runtime of EVARS-GPR averaged over all datasets was more than six times lower. It is not surprising that these comparison partners outperformed EVARS-GPR with respect to predictive power, however at the cost of computational runtime. Regarding *AirPassengers* and *MaunaLoa*, for which EVARS-GPR was outperformed in terms of RMSE, the CPU time of EVARS-GPR was 16 respectively 250 times more efficient. However, for *CashierData*, *DrugSales* and *VisitorNights*, the RMSE of EVARS-GPR was comparable to the leading ones, while being computationally much more efficient.

**Table 2. RMSE and CPU time compared to computationally exhaustive methods for datasets with output scale changes.** Numbers show the RMSE and the CPU time averaged over ten runs. The best RMSE results and lowest CPU times excluding $M_{base}$ are printed bold.

| | | $M_{base}$ | *PR1* | *PR2* | *MWGPR* | *EVARS-GPR* |
|---|---|---|---|---|---|---|
| *CashierData* | RMSE | 1351.43 | 1119.23 | 1185.82 | **1098.16** | 1125.34 |
| | CPU time [s] | 2.38 | 1443.80 | 741.71 | 1106.53 | **166.06** |
| *DrugSales* | RMSE | 6.15 | **2.44** | 2.54 | 2.86 | 2.75 |
| | CPU time [s] | 0.61 | 762.05 | 404.47 | 550.99 | **52.40** |
| *AirPassengers* | RMSE | 171.58 | **69.06** | 74.28 | 72.27 | 93.88 |
| | CPU time [s] | 0.93 | 587.70 | 303.31 | 514.53 | **34.53** |
| *MaunaLoa* | RMSE | 34.37 | 11.12 | 12.60 | **9.88** | 27.96 |
| | CPU time [s] | 3.60 | 27459.79 | 13790.61 | 19525.85 | **78.99** |
| *VisitorNights* | RMSE | 10.97 | **5.08** | 5.21 | 5.85 | 5.11 |
| | CPU time [s] | 0.35 | 40.85 | 22.28 | 33.87 | **6.24** |

In Table 3, we see that EVARS-GPR was robust for datasets without an output scale change during the online phase as the results were identical to $M_{base}$.

**Table 3. Robustness on datasets without output scale changes during the online phase.** Numbers show the RMSE of the simulated online phase.

| | *ChampagneSales* | *TouristsIndia* | *Milk* | *Beer* | *USDeaths* |
|---|---|---|---|---|---|
| $M_{base}$ | 1158.26 | 90707.30 | 15.16 | 16.88 | 276.72 |
| *EVARS-GPR* | 1158.26 | 90707.30 | 15.16 | 16.88 | 276.72 |

### 6.3 Discussion

We showed that EVARS-GPR is able to handle seasonal time series with changes of the target variable scale, both on simulated and real-world data. The performance on simulated data demonstrates a broad applicability regarding the time of occurrence, duration, speed and extent of the output scale change, with a higher advantage over $M_{base}$ for longer durations. Shorter changes are more difficult to detect for online CPD methods, which is one reason for the lower improvement in such settings. Our results further indicate that EVARS-GPR can handle various speeds and extents of the output scale change, which can be seen as different abruptness levels. This applies both for increases as well as decreases. Experiments with smaller extents showed smaller improvements of EVARS-GPR, as it is more difficult to detect such changes. A similar effect can be observed for smaller speeds and larger extents of the output scale change. Nevertheless, EVARS-GPR was at least on par with $M_{base}$ in all cases and outperformed it in most of the settings.

In addition, EVARS-GPR outperformed all methods with a similar computational resource consumption with respect to RMSE on real-world datasets, with a mean improvement of 20.8 % compared to the second-best approaches. Regarding *AirPassengers* and *MaunaLoa*, the advantage of EVARS-GPR in terms of RMSE was 7.6 % respectively 10.4 %. For these datasets, the output scale changes at the detected change points were rather small. Consequently, the DA step did not enhance the performance that much, which might be a reason for the smaller improvements on RMSE in contrast to the other datasets. Furthermore, the difference to other periodical refitting strategies was largest for *AirPassengers* and *MaunaLoa*. This might indicate that not all possible change points were detected or that these datasets possess further data distribution shifts not resulting in an output scale change. With respect to the other three datasets, EVARS-GPR's results were comparable to the periodical refitting strategies, suggesting that all relevant change points could be detected. Moreover, we showed EVARS-GPR's efficiency in comparison with periodical refitting strategies with a more than six-fold reduction of the averaged runtime in relation to *PR2*. This advantage of EVARS-GPR was even bigger for *AirPassengers* and *MaunaLoa* with a 16 respectively 250 times lower runtime compared to the best performer. Finally, EVARS-GPR was robust for datasets without changes of the target variable scale.

The online detection of change points is an essential part of EVARS-GPR, as wrong or missed detections might lead to a performance decrease. We addressed the problem of misdetections due to outliers with the introduction of a threshold for the output scaling factor. Nevertheless, EVARS-GPR would probably benefit a lot from improvements of the online CPD method. Moreover, we observed lower RMSE values for periodical refitting strategies, especially on a dataset with more samples (*MaunaLoa*). Thus, a combination of EVARS-GPR and a periodical refitting strategy with a lower frequency is an interesting approach for future research. This might result in a computationally efficient algorithm, which is additionally able to capture changing data distributions that do not result in a target variable scale. We further determined the parameters of the whole pipeline based on simulated data. This might not be the best strategy for all settings and real-world applications. However, as the simulated scenarios were diverse and reflected the scope of this work with output scale changes, this is a reasonable approach. Nevertheless, another way for parameter optimization is a further potential for future research. One possibility is to integrate this into the cross-validation performed offline by simulating different manipulations of the real-world data. Beyond that, EVARS-GPR is model-agnostic. Therefore, it seems interesting to transfer this approach to other prediction models, e.g. XGBoost, which is limited to prediction values within the target value range of the training set [8].

## 7 Conclusion

In this paper, we presented EVARS-GPR, a novel online time series forecasting algorithm that is able to handle sudden shifts in the target variable scale of seasonal data by combining change point monitoring with an augmented refitting of a prediction model. Online change point detection and data augmentation are essential components of EVARS-GPR, for which we evaluated different approaches based on simulated scenarios. Using the resulting configuration and parameterization, we showed on simulated data that EVARS-GPR is applicable for a wide range of output scale changes. Furthermore, EVARS-GPR had on average a 20.8 % lower RMSE on different real-world datasets compared to methods with a similar computational resource consumption. Moreover, we demonstrated its computational efficiency compared to periodical refitting strategies with a more than six-fold reduction of the averaged runtime.

**Acknowledgments.** This work is supported by funds of the Federal Ministry of Food and Agriculture (BMEL) based on a decision of the Parliament of the Federal Republic of Germany via the Federal Office for Agriculture and Food (BLE) under the innovation support program [grant number 2818504A18]. Furthermore, we acknowledge Maura John and Nikita Genze for fruitful discussions regarding the naming of the algorithm.

**Competing interests.** All authors declare that they have no competing interests.

## Appendix 1: Gaussian Process Regression

With regard to the practical use of forecasts, the uncertainty of a prediction value seems profitable. Providing those by its definition is a main advantage of the nonparametric Bayesian method GPR. To explain this approach, we use the linear model that is defined as

$$f(\boldsymbol{x}) = \boldsymbol{x}^T \boldsymbol{w}, \quad y = f(\boldsymbol{x}) + \epsilon \tag{2}$$

with $\boldsymbol{x}$ being the input vector, $\boldsymbol{w}$ the vector of weights, the function value $f(\boldsymbol{x})$ and observed target value $y$ with additive noise $\epsilon$ assumed to follow a zero-mean Gaussian. Combined with the independence assumption of the observation values, we get the likelihood, which reflects how probable the observed target values $\boldsymbol{y}$ are for the different inputs $\boldsymbol{X}$ and weights $\boldsymbol{w}$:

$$p(\boldsymbol{y}|\boldsymbol{X}, \boldsymbol{w}) = \prod_{i=1}^{j} p(y_i|\boldsymbol{x}_i, \boldsymbol{w}) \tag{3}$$

As usual for a Bayesian formulation, we define a prior over the weights, for which we again choose a zero-mean Gaussian. With the defined prior and the likelihood based on the observed data, we can use Bayes' rule to get the posterior of the weights given the data:

$$p(\boldsymbol{w}|\boldsymbol{X}, \boldsymbol{y}) = \frac{p(\boldsymbol{y}|\boldsymbol{X}, \boldsymbol{w})p(\boldsymbol{w})}{p(\boldsymbol{y}|\boldsymbol{X})} \tag{4}$$

This is also called the maximum a posteriori estimate, which - provided the data - delivers the most likely set of weights $\boldsymbol{w}$. As $p(\boldsymbol{y}|\boldsymbol{X})$ is independent of $\boldsymbol{w}$, we can reformulate this equation expressing the posterior distribution with a Gaussian defined by a mean and covariance matrix:

$$p(\boldsymbol{w}|\boldsymbol{X}, \boldsymbol{y}) \sim \mathcal{N}(\overline{\boldsymbol{w}}, \boldsymbol{A}^{-1}) \tag{5}$$

During inference, we marginalize out $\boldsymbol{w}$ and as a result take the average based on all possible $\boldsymbol{w}$ weighted by their posterior probability:

$$\begin{aligned} p(y_{Test}|\boldsymbol{x}_{Test}, \boldsymbol{X}, \boldsymbol{y}) &= \int p(y_{Test}|\boldsymbol{x}_{Test}, \boldsymbol{w})\, p(\boldsymbol{w}|\boldsymbol{X}, \boldsymbol{y})\, d\boldsymbol{w} \\ &= \mathcal{N}\left(\frac{1}{\sigma^2}\, \boldsymbol{x}_{Test}^T\, \boldsymbol{A}^{-1}\, \boldsymbol{X}\, \boldsymbol{y},\ \boldsymbol{x}_{Test}^T\, \boldsymbol{A}^{-1}\, \boldsymbol{x}_{Test}\right) \end{aligned} \tag{6}$$

Therefore, we do not only get an output value, but also an uncertainty. So far, we reached the Bayesian formulation of linear regression with its limited expressiveness. To overcome this constraint to linearity, we can project the inputs into a high-dimensional space and apply the linear concept there. This transformation can be accomplished using basis functions $\phi(\boldsymbol{x})$: $\mathbb{R}^d \to \mathbb{R}^i$ leading to the following model with $i$ weights $\boldsymbol{w}$:

$$f(\boldsymbol{x}) = \phi(\boldsymbol{x})^T \boldsymbol{w} \tag{7}$$

Conducting the same derivation as shown above results in a similar outcome:

$$p(y_{Test}|\boldsymbol{x}_{Test}, \boldsymbol{X}, \boldsymbol{y}) = \mathcal{N}\left(\frac{1}{\sigma^2}\phi(\boldsymbol{x}_{Test})^T \boldsymbol{A}^{-1}\, \Phi(\boldsymbol{X})\boldsymbol{y}, \phi(\boldsymbol{x}_{Test})^T \boldsymbol{A}^{-1}\, \phi(\boldsymbol{x}_{Test})\right) \tag{8}$$

The need of inverting the $ixi$ matrix $\boldsymbol{A}$ possibly causes computational problems if the dimension of the feature space $i$ becomes large. To solve this, we can reformulate the above using the so-called "kernel trick". This leads to the formulation of a Gaussian Process, which is completely specified by its mean and covariance function:

$$f(\boldsymbol{x}) \sim GP\big(m(\boldsymbol{x}), k(\boldsymbol{x}, \boldsymbol{x}')\big) \tag{9}$$

$$m(\boldsymbol{x}) = \mathbb{E}[\boldsymbol{f}(\boldsymbol{x})] \tag{10}$$

$$\boldsymbol{k}(\boldsymbol{x}, \boldsymbol{x}') = \mathbb{E}\big[\big(\boldsymbol{f}(\boldsymbol{x}) - \boldsymbol{m}(\boldsymbol{x})\big)\big(\boldsymbol{f}(\boldsymbol{x}') - \boldsymbol{m}(\boldsymbol{x}')\big)\big] \tag{11}$$

$k(\boldsymbol{x}, \boldsymbol{x}')$ consists of the covariance value between any two sample points $\boldsymbol{x}$ and $\boldsymbol{x}'$ resulting in a $nxn$ matrix for a training set length of $n$. The assumption is that the similarity between samples reflects the strength of the correlation between their corresponding target values. Therefore, the function evaluation can be seen as a draw from a multivariate Gaussian distribution defined by $m(\boldsymbol{x})$ and $k(\boldsymbol{x}, \boldsymbol{x}')$. Thus, Gaussian Processes are a distribution over functions rather than parameters, in contrast to Bayesian linear regression. For simplicity, the mean function is often set to zero or a constant value. There are many forms of kernel functions, which need to fulfill certain properties, e.g. being positive semidefinite and symmetric. Furthermore, they can be combined, e.g. by summation or multiplication. The choice of the covariance kernel function is a determining configuration of GPR and its parameters need to be optimized during training [15, 17].

# Appendix 2: List of Symbols

## General Symbols

| | |
|---|---|
| $M$ | prediction model |
| $t$ | current time step |
| $\mathcal{D}$ | time series dataset |
| $n$ | number of samples of the time series dataset $\mathcal{D}$ |
| $n_{off}$ | number of samples that are available during the offline phase |
| $\boldsymbol{x}_t$ | covariate vector at time step $t$ |
| $d$ | dimensionality of the covariate vector $\boldsymbol{x}_t$ |
| $\boldsymbol{X}$ | covariate matrix including all covariates vectors |
| $\boldsymbol{\chi}_t$ | subset of the covariate vector $\boldsymbol{x}_t$ at time step $t$ |
| $\boldsymbol{s}_{\chi,t}$ | seasonal component of the subset of the covariate vector $\boldsymbol{\chi}_t$ at time step $t$ |
| $\boldsymbol{r}_{\chi,t}$ | residual component of the subset of the covariate vector $\boldsymbol{\chi}_t$ at time step $t$ |
| $y_t$ | true target value at time step $t$ |
| $s_t$ | seasonal component of $y_t$ at time step $t$ |
| $r_t$ | residual component of $y_t$ at time step $t$ |
| $n_{seas}$ | length of one season |
| $\hat{y}_t$ | predicted target value at time step $t$ |
| $p(y\|\boldsymbol{x}_t)$ | generative distribution of $y$ |

## EVARS-GPR

| | |
|---|---|
| $M_{base}$ | offline-trained base model |
| $M_{current}$ | current prediction model |
| $\eta$ | current output scaling factor |
| $\eta_{old}$ | output scaling factor of last augmented refitting |
| $n_w$ | window size for the calculation of the output scaling factor $\eta$ |
| $n_\eta$ | number of previous seasons considered for the calculation of the output scaling factor $\eta$ |
| $\pi_\eta$ | minimum threshold for the deviation between the current $\eta$ and the last output scaling factor $\eta_{old}$ |
| $\mathcal{D}'$ | augmented set of samples |

## Simulated Data

| | |
|---|---|
| $a$ | amplitude |
| $n_{seas}$ | length of a season |
| $t_{start}$ | start index of the output scale change |
| $t_{end}$ | end index of the output scale change |
| $\delta$ | multiplicative manipulation factor for the output scale change |
| $\delta_{base}$ | starting manipulation factor |
| $\delta_{max}$ | maximum manipulation factor for the output scale change |
| $\kappa$ | slope, i.e. increase per time step $t$, for the manipulation factor $\delta$ |

## Appendix 3: Online Change Point Detection

The goal of CPD is to find abrupt changes in data, in the context of this work resulting in a shift of the scale of the target variable $y$. Based on the criteria outlined in section 4, we selected Bayesian Online Change Point Detection and ChangeFinder.

Bayesian Online Change Point Detection (BOCPD) is a common probabilistic technique. BOCPD assumes that a sequence of observations $y_1, y_2, \dots, y_T$ can be divided into non-overlapping partitions $\rho$ within which the data is i.i.d. from a distribution $p(y_t|\theta_\rho)$, with the parameters $\theta_\rho$ being i.i.d. as well. A central aspect of BOCPD is the definition of the run length at time step $t$, $r_t$, i.e. the time since the last change point. The posterior distribution of the run length $r_t$ can be determined using Bayes' theorem with $y_t^{(r)}$ denoting the observations associated with $r_t$:

$$p(r_t|y_{1:t}) = \frac{\sum_{r_{t-1}} p(r_t|r_{t-1})\, p(y_t|r_{t-1}, y_t^{(r)})\, p(r_{t-1}, y_{1:t-1})}{\sum_{r_t} p(r_t, y_{1:t})} \tag{12}$$

The conditional prior $p(r_t|r_{t-1})$ is defined to be nonzero only for $r_t = 0$ and $r_t = r_{t-1} + 1$ making the algorithm computationally efficient:

$$p(r_t|r_{t-1}) = \begin{cases} H(r_{t-1}+1) & if\ r_t = 0 \\ 1 - H(r_{t-1}+1) & if r_t = r_{t-1} + 1 \\ 0 & otherwise \end{cases} \tag{13}$$

$H(\tau) = \frac{p_{gap}(g=\tau)}{\sum_{t=\tau}^{\infty} p_{gap}(g=t)}$ is the so-called hazard function with the *a priori* probability distribution over the interval between between change points $p_{gap}(g)$. To apply BOCPD, a hazard function needs to be provided. With $p_{gap}(g)$ being a discrete exponential distribution with timescale $\lambda$, the hazard function is constant at $H(\tau) = 1/\lambda$, which is a common assumption. Finally, using the posterior distribution of the run length $r_t$, a change point can be determined [1, 2].

Another type of online CPD techniques suitable for our purposes are likelihood ratio methods, which declare a change point if the probability distributions before and after a candidate point differ significantly. ChangeFinder is a common approach of this kind, which employs a two-stage learning and smoothing strategy using Sequentially Discounting Auto-Regression (SDAR) model learning. In the first stage, we fit an SDAR model for each new sample at time step $t$ to represent the statistical behavior of the data. The model parameters are updated sequentially with a reduction of the influence of older samples. Thereby, we obtain a sequence of probability densities $p_1, p_2, \dots, p_t$ for each $y_t$. Based on these, we assign an outlier score to each data point, which is defined as $score(y_t) = -\log p_{t-1}(y_t)$. This enables the formulation of an auxiliary time series $o_t$ by building moving averages of the outlier scores within a time window $T$ for each time step $t$:

$$o_t = \frac{1}{T} \sum_{i=t-T+1}^{t} score(y_i) \tag{14}$$

After this first smoothing, the second stage starts. Thereby, another SDAR model is fitted using $o_t$, also resulting in a sequence of probability densities $q_1, q_2, \ldots, q_t$. Finally, we get a change point score $z_t$ after a second smoothing step within a time window $T$:

$$z_t = \frac{1}{T}\sum_{i=t-T+1}^{t} -\log q_{t-1}(o_t) \quad (15)$$

A higher value of $z_t$ corresponds to a higher probability of a change point at time step $t$. Hence, a threshold $\pi_{cf}$ at which a change point is declared, needs to be defined [2, 14, 25].

## Appendix 4: Data Augmentation

As outlined in section 4, we used Virtual Sample Generation approaches for imbalanced regression besides a scaling of the original dataset for data augmentation. These methods are suitable for continuous target variables and small datasets. We therefore selected the two following approaches: Random Oversampling with the introduction of Gaussian Noise (GN) [26] and SMOGN, which combines the former and the Synthetic Minority Oversampling TEchnique for Regression (SMOTER) [5, 27]. Both methods start with the assignment of a relevance value to every sample $(\boldsymbol{x}_i, y_i)$ of a dataset $\mathcal{D}_{vsg}$ using a relevance function $\phi: y \rightarrow [0,1]$. Based on these and a specified threshold $\pi_{rel}$, the dataset $\mathcal{D}_{vsg}$ is splitted into a subset of normal and rare cases, $\mathcal{D}_N$ respectively $\mathcal{D}_R$. We employed a relevance function, which proposes an inverse proportionality of the relevance value and the probability density function of $y$ [23]. Therefore, extreme cases have a higher relevance value. Furthermore, we tested two compositions of $\mathcal{D}_{vsg}$. In the first case, $\mathcal{D}_{vsg}$ was equal to the original dataset up to the change point at time step $t$, $\mathcal{D}_{0:t}$, and in the second one $\mathcal{D}_{0:t}$ and the output scaled dataset $\mathcal{D}_{0:t}^{\eta}$ were concatenated. Both GN and SMOGN then apply a Random Undersampling strategy for the normal cases $\mathcal{D}_N$, meaning that a specified share of these is randomly selected to get $\mathcal{D}_{us}$.

Furthermore, GN generates new samples $\mathcal{D}_{os}$ based on the rare cases $\mathcal{D}_R$ by adding Gaussian Noise to the target variable as well as the numeric covariates. Values for nominal attributes are randomly selected with a probability proportional to their frequency in the dataset. Finally, for GN, the undersampled and oversampled cases are concatenated to the augmented dataset $\mathcal{D}^{GN} = \{\mathcal{D}_{us}, \mathcal{D}_{os}\}$.

SMOGN instead employs two different oversampling techniques: GN and SMOTER. Prior to the sample generation, the k-nearest neighbors of a seed sample are determined. If a randomly selected k-nearest neighbor is within a specified maximum distance, SMOTER is performed resulting in a set of new samples $\mathcal{D}_{os}^{SMOTER}$. With SMOTER, values of numeric attributes are interpolated and nominal ones are randomly selected. The target value is determined by a weighted average with weights that are inversely proportional to the distance between the seed samples and the new generated one. In case the maximum distance is exceeded, GN is performed, leading to a second oversampling dataset $\mathcal{D}_{os}^{GN}$. The final augmented dataset for SMOGN therefore consists of three sets: $\mathcal{D}^{SMOGN} = \{\mathcal{D}_{us}, \mathcal{D}_{os}^{SMOTER}, \mathcal{D}_{os}^{GN}\}$ [5, 6, 23].

## Appendix 5: EVARS-GPR parameters

**Table 4.** Overview of EVARS-GPR's parameters with all analyzed methods CPD and DA.

| category | parameter | explanation |
|---|---|---|
| general parameters | scale_window*(_factor)* | size of window prior to detected change point for calculation of output scaling factor, alternatively formulated as a factor of the season length |
| | scale_window_minimum | minimum window size for output scaling factor |
| | scale_seasons | number of seasons considered for output scaling factor |
| | scale_thr | minimum threshold for the deviation between the current and last output scaling factor to trigger augmented refitting |
| CPD parameters | ***BOCPD*** | |
| | const_hazard*(_factor)* | constant of the hazard function, alternatively formulated as a factor of the season length |
| | ***ChangeFinder*** | |
| | cf_r | forgetting factor of the SDAR models |
| | cf_order | order of the SDAR models |
| | cf_smooth | window size for the smoothing step |
| | cf_thr_perc | percentile threshold of the anomaly score during the offline phase to declare a change point |
| DA parameters | max_samples*(_factor)* | maximum number of samples for DA, alternatively formulated as number of seasons |
| | ***GN*** | |
| | gn_operc | oversampling percentage |
| | gn_uperc | undersampling percentage |
| | gn_thr | threshold to determine normal and rare values |
| | append | specify if scaled dataset version is appended prior to sample generation |
| | ***SMOGN*** | |
| | smogn_relthr | threshold to determine normal and rare values |
| | smogn_relcoef | box plot coefficient for relevance function |
| | smogn_under_samp | specify if undersampling is performed |
| | append | specify if scaled dataset version is appended prior to sample generation |

Based on the results over all 67 simulated scenarios, we selected CF for online CPD and a scaling of the original dataset for DA. Furthermore, our experiments yielded the following parameters: scale_window_factor = 0.1, scale_window_minimum = 2, scale_seasons = 2, scale_thr = 0.1, cf_r = 0.4, cf_order = 1, cf_smooth = 4, cf_thr_perc = 70. For efficiency, we set max_samples_factor = 10.

## Appendix 6: Real-World Datasets

**Table 5.** Overview of the used real-world datasets.

| dataset | explanation | samples |
|---|---|---|
| ***datasets with a changing output scale during the online phase*** | | |
| *CashierData* | weekly sales of a horticultural retailer | 195 |
| *DrugSales* | sales of antidiabetic drugs per month | 204 |
| *VisitorNights* | visitor nights per quarter in millions in Australia | 68 |
| *AirPassengers* | total number of US airline passengers per month | 144 |
| *MaunaLoa* | monthly averaged parts per million of $CO_2$ measured at Mauna Loa observatory, Hawaii | 751 |
| ***datasets without a changing output scale during the online phase*** | | |
| *ChampagneSales* | sales of perrin freres champagne | 105 |
| *TouristsIndia* | foreign tourist arrivals per quarter in India | 48 |
| *Milk* | average milk production per cow and month | 168 |
| *Beer* | Australian monthly beer production | 56 |
| *USDeaths* | accidental deaths in the US per month | 72 |

**Table 6.** Overview of additional calendric and statistical features. Not all features are applicable for all datasets, e.g. due to the temporal resolution. Features are added to existing ones.

| category | features | explanation |
|---|---|---|
| calendric features | date based features | hour, day of month, weekday, month, quarter |
| | working day | flag showing if the day is a working day |
| statistical features | lagged variables | prior values of the target variable / features |
| | seasonal lagged variables | prior values of the preceding season |
| | rolling statistics | rolling mean and maximum within a window |
| | seasonal rolling statistics | seasonal rolling mean and maximum within a window |
| | rolling weekday statistics | rolling mean and maximum within a window calculated for each weekday |

# Appendix 7: Further Simulated Scenarios

The following figures show further simulated scenarios. Each box shows the result of the scenario parametrized with the value given on the x- and y- axis. Numbers are RMSE-ratios, lower values reflect a higher improvement compared to $M_{base}$.

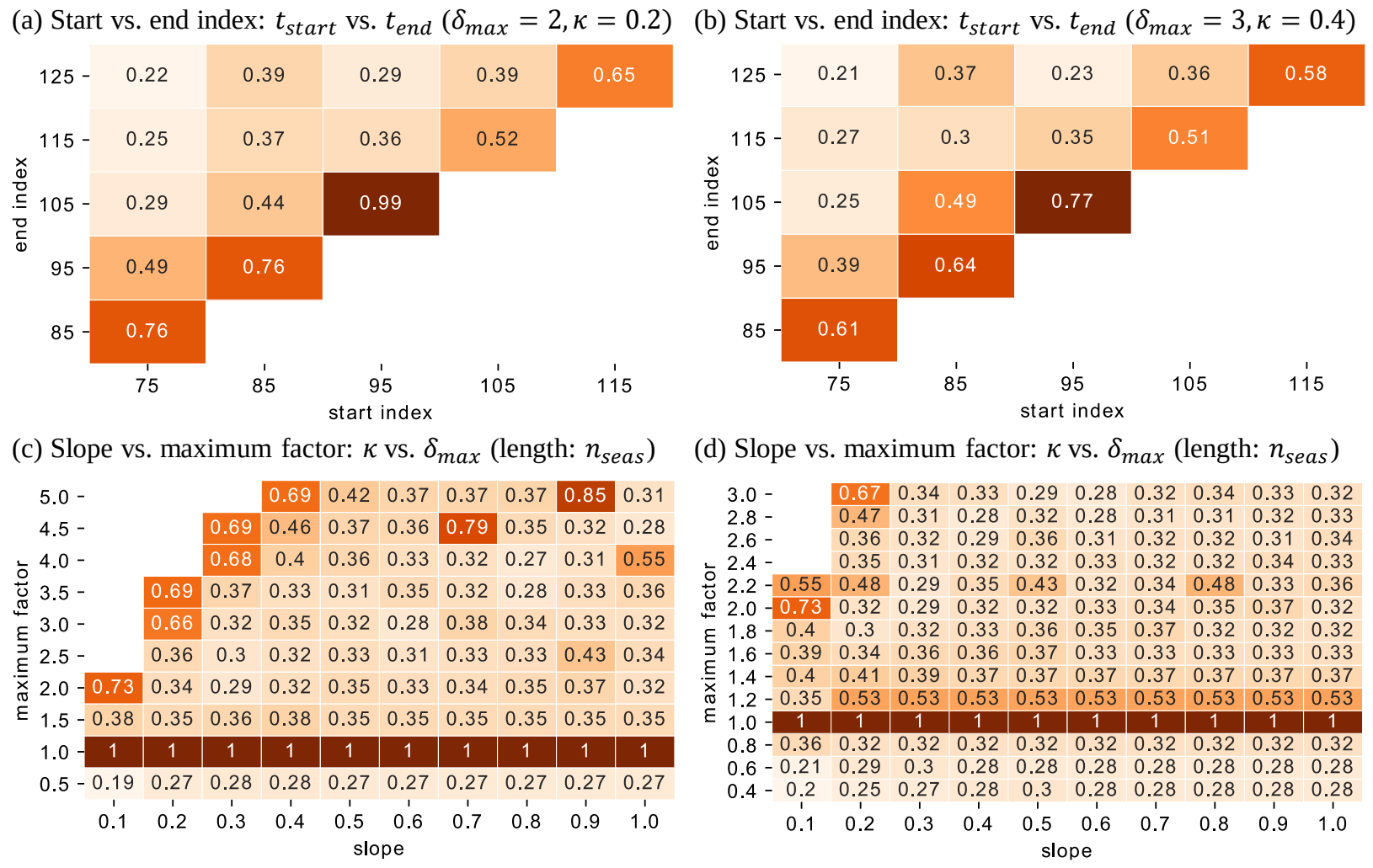

**Fig. 4. Behavior on a variety of simulated data ($n_{seas} = 25$). (a)** time of occurrence and duration via start and end indices $t_{start}$ respectively $t_{end}$, **(b)** same factors with a higher slope $\kappa$ and maximum manipulation factor $\delta_{max}$, **(c)** extent and speed of the change via $\kappa$ and $\delta_{max}$, **(d)** same factors on a finer grid for $\delta_{max}$.

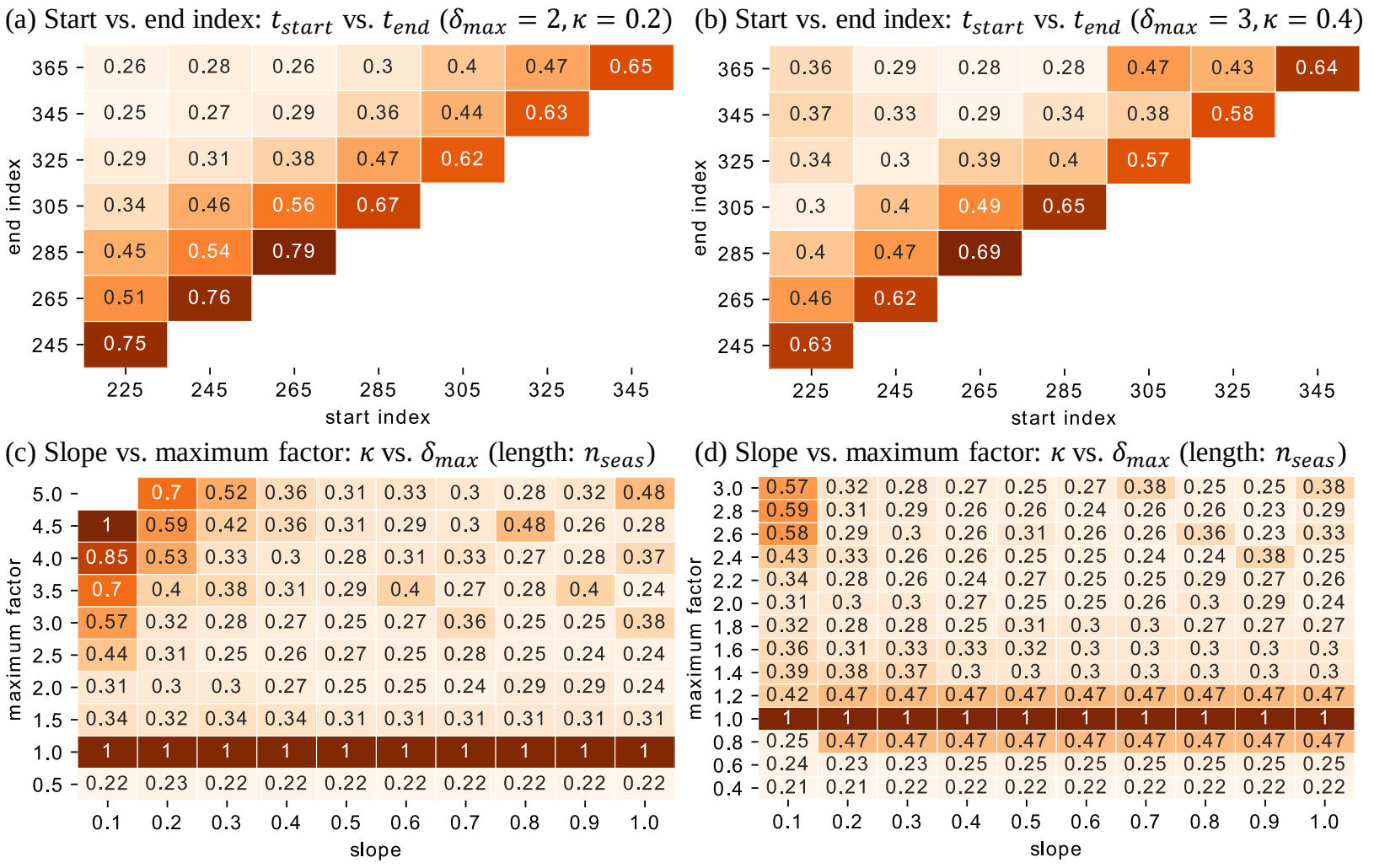

**Fig. 5. Behavior on a variety of simulated data ($n_{seas} = 75$). (a)** time of occurrence and duration via start and end indices $t_{start}$ respectively $t_{end}$, **(b)** same factors with a higher slope $\kappa$ and maximum manipulation factor $\delta_{max}$, **(c)** extent and speed of the change via $\kappa$ and $\delta_{max}$, **(d)** same factors on a finer grid for $\delta_{max}$.

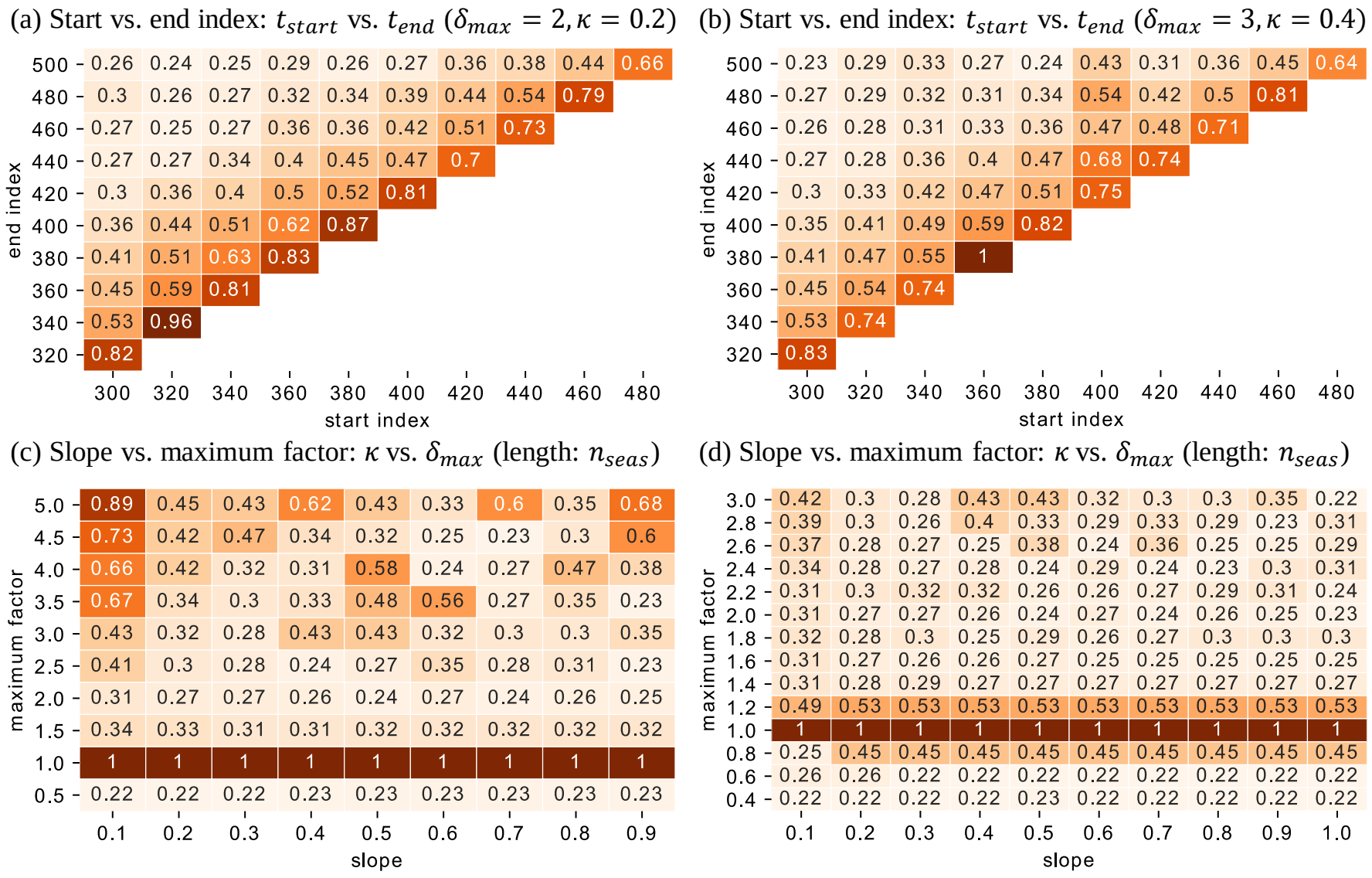

**Fig. 6. Behavior on a variety of simulated data ($n_{seas} = 100$). (a)** time of occurrence and duration via start and end indices $t_{start}$ respectively $t_{end}$, **(b)** same factors with a higher slope $\kappa$ and maximum manipulation factor $\delta_{\max}$, **(c)** extent and speed of the change via $\kappa$ and $\delta_{max}$, **(d)** same factors on a finer grid for $\delta_{max}$.